\documentclass[10pt,twocolumn,letterpaper]{article}

\usepackage[pagenumbers]{iccv} 

\definecolor{iccvblue}{rgb}{0.21,0.49,0.74}
\usepackage[pagebackref,breaklinks,colorlinks,allcolors=iccvblue]{hyperref}
\usepackage{multirow}
\usepackage{array}
\usepackage{graphicx}

\newcolumntype{P}[1]{>{\centering\arraybackslash}p{#1}}

\def\paperID{12778} 
\def\confName{ICCV}
\def\confYear{2025}

\title{SuperCap: Multi-resolution Superpixel-based Image Captioning}

\author{Henry Senior\\
Queen Mary University of London\\
London, UK\\
{\tt\small h.senior@qmul.ac.uk}
\and
Luca Rossi\\
The Hong Kong Polytechnic University\\
Hong Kong\\
{\tt\small luca.rossi@polyu.edu.hk}
\and
Gregory Slabaugh\\
Queen Mary University of London\\
London, UK\\
{\tt\small g.slabaugh@qmul.ac.uk}
\and
Shanxin Yuan\\
Queen Mary University of London\\
London, UK\\
{\tt\small shanxin.yuan@qmul.ac.uk}
}

\begin{document}
\maketitle
\begin{abstract}
It has been a longstanding goal within image captioning to move beyond a dependence on object detection. We investigate using superpixels coupled with Vision Language Models (VLMs) to bridge the gap between detector-based captioning architectures and those that solely pretrain on large datasets. Our novel superpixel approach ensures that the model receives object-like features whilst the use of VLMs provides our model with open set object understanding. Furthermore, we extend our architecture to make use of multi-resolution inputs, allowing our model to view images in different levels of detail, and use an attention mechanism to determine which parts are most relevant to the caption. We demonstrate our model's performance with multiple VLMs and through a range of ablations detailing the impact of different architectural choices. Our full model achieves a competitive CIDEr score of $136.9$ on the COCO Karpathy split.
\end{abstract}    
\section{Introduction}
\label{sec:intro}

Image captioning, the challenging task of developing a model that describes a visual scene in natural language, has received considerable attention in recent years from the computer vision community. It is a task with great scientific impact (\eg, ensuring models can truly describe the scene) and societal impact (\eg, captioning models provide benefits in terms of accessibility through generating alt-text for websites). While early works focused on a standard Encoder-Decoder architecture~\cite{karpathy2015deep} using a Convolutional Neural Network (CNN) to produce a feature vector that is decoded into a caption via a Recurrent Neural Network (RNN), later works moved to adopt object detectors.

There are typically two approaches to image captioning: detector-based (Bottom Up Top Down (BUTD) \cite{anderson2018bottom}  and its descendants~\cite{cornia2020meshed, zhong2020comprehensive, zhao2019object}) or `detector-free' models, the latter of which are mostly massively pretrained large Vision-Language Models (VLMs) built on top of the Transformer architecture~\cite{vaswani2017attention} and pretrained on large datasets such as LAION-5B~\cite{schuhmann2022laion}. These models produce strong global features but risk missing out on more nuanced finer grain detail, something at which detector-based methods naturally excel. In this work, we investigate taking the best of both worlds, answering the question, `how can we leverage VLMs to produce high quality captions whilst utilising finer object details?'. We achieve this through the use of superpixels: by segmenting the image into $M$ superpixels, we leverage the zero shot classification capabilities of VLMs to replace the object detector. To ensure finer-details are captured, we incorporate a multi-resolution approach, where the image is divided into a set of superpixels of different resolutions. We also make use of a whole image feature to maintain the global context.

Early deep learning approaches to image captioning focused on using a CNN to represent the entire image using a unique latent feature. This feature was then decoded into a caption using an RNN. Although this approach was initially successful, it produced captioning models that would focus on the global context of an image and would therefore miss specific objects~\cite{karpathy2015deep}. As object detectors improved, they began to be used in image captioning models to address the weaknesses of earlier approaches. Following on from Anderson \textit{et al.}'s BUTD \cite{anderson2018bottom} work, a large number of image captioning architectures based on object detectors have been produced ~\cite{cornia2020meshed, zhong2020comprehensive, zhao2019object}. Whilst these models all provide strong performance and novel ways of utilising the objects in different ways, they all share a common flaw - they are fundamentally limited by their object detectors. Specifically, 1) the closed set nature of object detectors, which results in only objects in the training data of the detector being identified, and 2) the expensive computation required by object detectors (FasterRCNN \cite{ren2015faster} with a ResNet-50 \cite{he2016deep} takes 447 GFLOPs). As object detectors are only as good as their training data, in scenarios where the detector misclassifies an object (or is unable to classify it), the caption will be incorrect.

The closed-set nature of detector-based image captioning, alongside the development of large pretrained VLMs has led researchers to develop detector-free captioning architectures~\cite{wang2022git}. These approaches largely address the problems of detector-based architectures by scaling the Transformer and the data used to train the model. Given the impressive Transformer scaling laws~\cite{kaplan2020scaling}, the development in computing infrastructure, and the increasing accessibility of ever larger datasets, this approach has proven to be the dominant approach in recent years. 

Methods aiming to unify detector-based and detector-free are limited~\cite{fang2022injecting}. Their architecture mirrors that of the detector-free models~\cite{wang2022git}, a ViT combined with a transformer language model. However, they include a novel `concept token network' that predicts concepts from the hidden layers of the ViT, essentially providing object detection via proxy. This approach leads to a complex model that is difficult to train. In comparison, our model is easily trainable end-to-end following standard procedures for an image captioning model trained on COCO~\cite{lin2014microsoft, chen2015microsoft}.

In this paper we present SuperCap, a novel detector-free image captioning architecture that joins together the benefits of detector-based and detector-free approaches to image captioning. The key contributions of our paper are as follows:
\begin{enumerate}
    \item SuperCap is the first image captioning model to be based on superpixels instead of patches or object bounding boxes. Our model performs competitively on standard benchmarks whilst having an architecture that scales appropriately to future developments within the computer vision community.
    \item To the best of our knowledge, we are the first image captioning architecture to make explicit use of multiple resolutions, thus ensuring that our model captures the local context (like detector-based models) and the global context (like pretrained detector-free models). We empirically show the advantage of our multi-resolution approach through extensive ablation studies. 
\end{enumerate}
The remainder of this paper is organised as follows. Section~\ref{sec:background} reviews the
related work, while Sections~\ref{sec:method} introduces SuperCap. In Section~\ref{sec:implementation} we discuss the implementation details of our model as well as of the training and evaluation protocol. We then conduct an extensive experimental evaluation in Section~\ref{sec:results} and we conclude the paper in Section~\ref{sec:conclusion}.
\section{Related Work}
\label{sec:background}

In this section we detail the related work and background to provide context for the literature that works sits alongside. We start by giving an overview of image captioning, followed by detector-free approaches to this problem. The section concludes with a brief overview of superpixels.

\begin{figure*}[!t]
    \centering
    \begin{subfigure}[b]{0.3\textwidth}
        \centering
        \includegraphics[width=\textwidth]{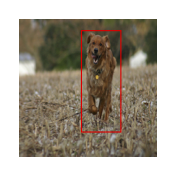}
        \subcaption{\textbf{Bounding Boxes} One common way in which regions are extracted is through the use of bounding boxes. This is used by Anderson \textit{et al.}~\cite{anderson2018bottom} and subsequent works~\cite{cornia2020meshed, yao2018exploring, zhong2020comprehensive}.}
        \label{fig:regions-bbox}
    \end{subfigure}
    \hfill
    \begin{subfigure}[b]{0.3\textwidth}
        \centering
        \includegraphics[width=\textwidth]{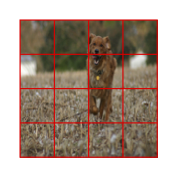}
        \subcaption{\textbf{Patches} A more contemporary approach to region extraction is through the use of patches~\cite{fang2022injecting}. However, it results in objects being divided between features, as shown above.}
        \label{fig:regions-patches}
    \end{subfigure}
    \hfill
    \begin{subfigure}[b]{0.3\textwidth}
        \centering
        \includegraphics[width=\textwidth]{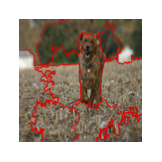}
        \subcaption{\textbf{Superpixels} Our approach utilises superpixels to create regions that encompass the whole image but ensure that objects are not unnecessarily divided between features.}
        \label{fig:regions-superpixels}
    \end{subfigure}
    \caption{\textbf{Visual Representation of Different Region Extraction Techniques.} Here we show the three different ways in which regions can be extracted from images, bounding boxes, patches, and superpixels.}
    \label{fig:regions}
\end{figure*}

\subsection{Image Captioning}
Given an input image $\mathcal{I}$, an image captioning model is tasked with producing a caption $\mathcal{C}=\{w_i\}^N_{i=1}$ comprised of $N$ words in a predetermined dictionary $w_i \in \mathcal{D}$. Most image captioning methods begin by defining a region extractor $\varphi$ which produces a set of $M$ regions $\mathcal{R}=\{\mathbf{r}_i\}^M_{i=1}$ detected within the image, \ie,
\begin{equation}
\varphi: \mathcal{I} \mapsto \mathcal{R}\,.
\end{equation}
Typically regions correspond to the bounding boxes of objects~\cite{anderson2018bottom, yao2018exploring, zhao2019object} or patches ~\cite{fang2022injecting}. These regions are then processed by a region feature extractor $\Psi$ which produces a set of latent features $\bar{\mathcal{{R}}} = \{\mathbf{r}_i \in \mathbb{R}^N\}_{i=1}^M$, \ie,
\begin{equation}
\Psi: \mathcal{R} \mapsto \bar{\mathcal{{R}}}\,.
\end{equation}
In the case of bounding boxes, $\Psi$ is typically a ResNet~\cite{he2016deep} model~\cite{anderson2018bottom, yao2018exploring, zhao2019object} or in the case of patches, $\Psi$ is comprised of a Vision Transformer~\cite{dosovitskiy2020image}.

Finally, the main captioning model $\Omega$ uses the latent features $\bar{\mathcal{{R}}}$ alongside the previously predicted word $w_{t-1}$ to predict the next word in $\mathcal{C}$, \ie,
\begin{equation}
\Omega: (\bar{\mathcal{{R}}}, w_{t-1}) \mapsto w_t \,.
\end{equation}
Traditionally~\cite{anderson2018bottom} this has been implemented by a dual LSTM but more contemporary approaches~\cite{fang2022injecting, wang2022git} make use of a Transformer.

Typically, $\varphi$, $\Psi$, and $\Omega$ are neural networks. We note $\varphi$ and $\Psi$ are not usually included in the end-to-end training of the captioning network, instead it is trained to complete a standalone task such as object detection.

Most models use the $\varphi$ and $\Omega$ defined in~\cite{anderson2018bottom}. That is, a Visual Genome \cite{krishna2017visual} based FasterRCNN~\cite{ren2015faster} object detector with a ResNet backbone~\cite{he2016deep} representing the region operators ($\varphi$ and $\Psi$) and a language model consisting of a dual LSTM with attention providing the captioning model ($\Omega$). The majority of methods (especially those in GNN-based image captioning~\cite{senior2023graph}) focus on augmenting the region encoder that produces rich latent features that capture the necessary information to produce high quality captions. Yao \etal \cite{yao2018exploring} implemented a GNN-based architecture that used both semantic and spatial graphs. Further approaches built on the foundation laid by Yao \etal \cite{yao2018exploring}, with Zhong \etal~\cite{zhong2020comprehensive} making use of a GCN~\cite{kipf2016semi} to perform semantic graph decomposition on the semantic graph. They address the problem of which nodes and edges of the graph to include in the final caption.

\subsection{Detector-Free Image Captioning}
One way in which research has addressed the issues introduced by object detectors is Vision-Language Pretraining (VLP), the practice of using massive amounts of data (typically from multiple datasets) to give the model a rich image-language embedding suitable for a large number of downstream tasks. Pretrained models are then finetuned on specific datasets, honing their performance at a particular task. This approach has been the main approach to detector-free image captioning~\cite{xu2021e2e, wang2022git, li2022blip} as large Vision-Language models learn to identify objects as an emergent behaviour learnt from the vast amounts of data.

More recently, Fang \etal~\cite{fang2022injecting} proposed a technique based on ViT~\cite{dosovitskiy2020image} that boasts strong performance without the need for VLP. Their approach uses a novel Concept Token Network, a Transformer-based~\cite{vaswani2017attention} network that takes the intermediary tokens of ViT~\cite{dosovitskiy2020image} as input, and predicts the top-$k$ most likely concept tokens from a corpus generated from COCO\cite{lin2014microsoft, chen2015microsoft}. These concept tokens are then concatenated with the final output feature of ViT to create a final latent feature vector that is decoded into a caption by a BERT-based~\cite{devlin2018bert} language model. Fang \etal's~\cite{fang2022injecting} Concept Token Network is essentially used to perform object detection via proxy and leads to a network architecture that requires an involved training strategy. Their reliance on large network architectures such as ViT and BERT results in an overall model with $190$ million parameters. Meanwhile, the architecture proposed in this paper sits at only $93$ million trainable parameters in its main configuration. 

Another approach to detector-free image captioning, which uses non-traditional region definitions (segmentation masks), is Segment and Caption Anything~\cite{Huang_2024_CVPR}. This model combines SAM~\cite{kirillov2023segment} and an LLM with a thin trainable network allowing for parameter efficient training of a region captioning model. However, to achieve strong performance, it requires large amounts of weak pretraining, something our model avoids via the use of VLMs. 

For a more detailed survey on image captioning, we direct the reader to Stefanini \etal ~\cite{stefanini2022show} and Senior \etal~\cite{senior2023graph}, the latter of which has a particular focus on GNN-based approaches.

\subsection{Superpixels}
Superpixel segmentation is a well established technique within computer vision for grouping pixels together into coherent and homogeneous boundaries. Unlike bounding boxes, superpixels are non-rigid and therefore easily handle edges of objects that are not straight. There are many approaches to producing superpixels for an image, including clustering, watershed, density, and graph-based ones, to name a few. One widely adopted approach is the Simple Linear Iterative Clustering (SLIC) proposed by Achanta \etal~\cite{achanta2010slic}. As a cluster-based approach, SLIC works by  converting the pixels into a $5$-dimensional space comprised of the $(L, a, b)$ pixel colour values of the CIELAB colour space and their $(x, y)$ location in the image. Once the pixels for the image are converted into this space, SLIC then clusters the values into $K$ groups, which then go on to become the $K$ superpixels. Although the SLIC superpixel algorithm appears simple at first glance, it provides an elegant and robust technique to provide the superpixel segmentation of a given image. 
\section{Method}
\label{sec:method}

In this section we introduce our architecture, both for the superpixel preprocessing and the captioning model itself. We start by outlining how our model extracts superpixel regions from a given image, following on with how we convert superpixels to region features which are used for captioning. Finally, we conclude by detailing how the architecture is extended to support multi-resolution.

\begin{figure*}[!th]
  \centering
   \includegraphics[width=\textwidth]{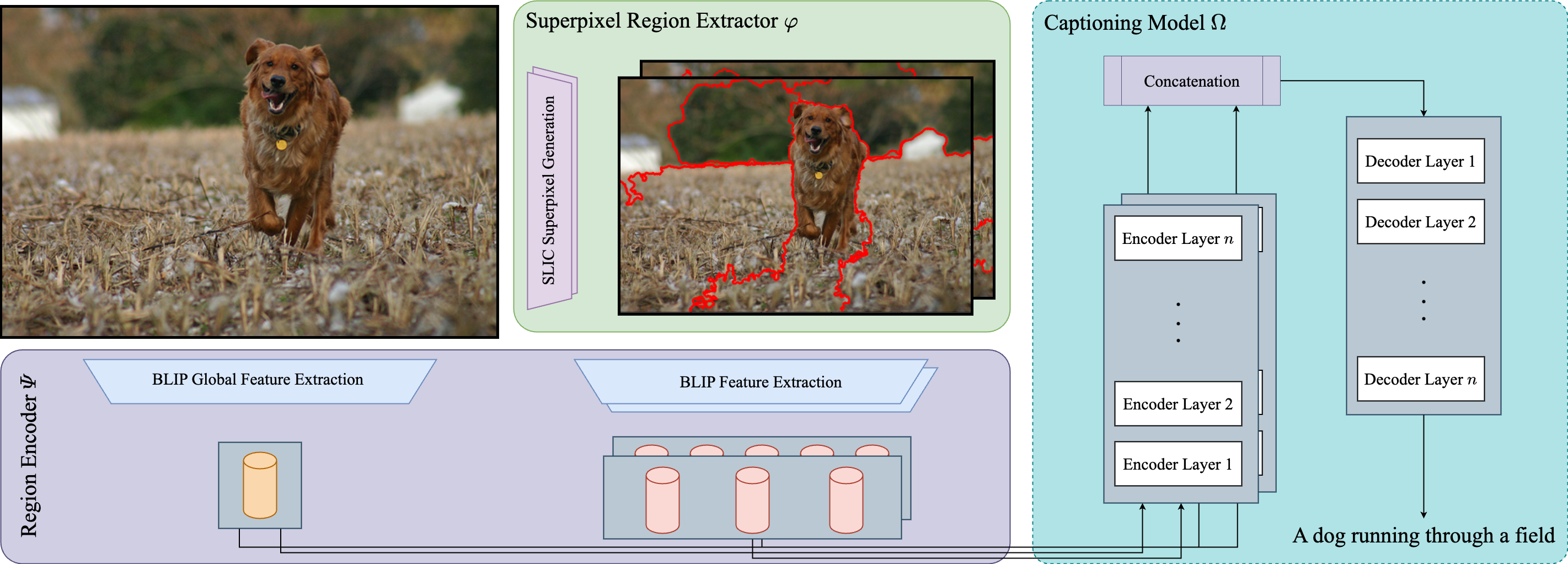}

   \caption{\textbf{Architecture Diagram} Our SuperCap model, like other image captioning models, has three main components: 1) the region extractor ($\varphi$), 2) the region encoder ($\Psi$), and 3) the captioning model ($\Omega$). The initial encoder takes in an image and uses BLIP to extract a global image feature. It simultaneously extracts superpixel regions for the image at different resolutions, which are in turn converted to BLIP region features. The secondary encoder consists of a stack of encoder transformers, one per resolution, which receive both the global feature and superpixel features for their resolution. Latent features produced by these encoders are then concatenated together and fed into a decoder transformer which autoregressively produces the final image caption.}
   \label{fig:architecture}
\end{figure*}

\subsection{Superpixel Regions Extraction ($\varphi$)}

The region extraction forms the first component of our architecture. Given an image, $\mathcal{I}$, it must produce a set of distinct regions $\mathcal{R}$ that can be used as input to a captioning architecture \ie $\mathcal{R} = \varphi(\mathcal{I})$. Whilst patches~\cite{fang2022injecting} or object detectors~\cite{anderson2018bottom} have been traditionally used for this task, we instead take the approach of using superpixels (shown in \Cref{fig:regions}). We choose to make use of superpixel regions as they produce contiguous groups of pixels that make up an object. This ensures that our regions are more homogenous and reduces the risk of objects being broken up unnecessarily. 

Specifically, we use the SLIC~\cite{achanta2010slic} superpixel segmentation technique for its computational efficiency~\cite{achanta2012slic}. Using SLIC, our model will produce $M_k$ superpixel regions, $\mathcal{R}_k = \{r_i\}_{i=1}^{M_k}$. To ensure that our model is able to capture different levels of image details, we extend the superpixel region extraction to include multiple resolutions. This results in $\mathcal{R} = \{\mathcal{R}_1, ..., \mathcal{R}_k\}$ regions being produced for $k$ resolutions. 

\subsection{Regions Encoder ($\Psi$)}

Once a set of regions have been defined, they must be converted into latent features, $\bar{\mathcal{R}}$, that can be used in the captioning model \ie $\bar{\mathcal{R}} = \Psi(\mathcal{R})$. To achieve this, we look towards VLMs such as CLIP~\cite{radford2021learning} and BLIP~\cite{li2022blip}. These large models produce features that combine semantic elements of both visual elements and language elements and are therefore a good fit for image captioning. We produce latent features for superpixels by generating a bounding box around each superpixel and feeding those extracted pixels into the VLM. This approach allows features to be generated for coherent regions of the image rather than smaller broken up regions, as is the case for patch based methods. For each resolution, $k$, the latent features are defined as
\begin{equation}
    \bar{\mathcal{{R}}_k} = \{\mathbf{r}_i \in \mathbb{R}^N\}_{i=1}^M, \hspace{1cm} \mathbf{r}_i = VLM(r_i)
\end{equation}
Additionally, we also introduce a global image feature $\mathbf{r}_0 \in \mathbb{R}^N$ to capture the wider context of the image.


\subsection{Captioning Model ($\Omega$)}
Once the superpixel region features ($\bar{\mathcal{R}}$) have been produced, the captioning model must refine them to better account for the wider context of the image. Our model implements this through an Encoder-Decoder transformer~\cite{vaswani2017attention} model. In order to support multiple resolutions, we stack $k$ Encoders, one per resolution, and concatenate their output before inputting the attended features into the decoder.

The encoder consists of a stack of a classic Transformer comprised of a multi-head self-attention blocks followed by a feed-forward network. The superpixel features for a given resolution $\bar{\mathcal{R}_k}$ are passed through the Transformer to calculate their corresponding attended features $\bar{\mathcal{R}_k}'$. This is achieved by first converting the input features $\mathcal{R}$ into corresponding queries, keys, and values by reprojecting them according to
\begin{equation}
    Q = \bar{\mathcal{R}_k}W_Q, \hspace{0.5cm} K = \bar{\mathcal{R}_k}W_K, \hspace{0.5cm} V = \bar{\mathcal{R}_k}W_V \,,
\end{equation}
where $W_Q, W_K, W_V$ are all learnt weight matrices. The attention weights ($\alpha$) are then calculated with
\begin{equation}
    \alpha = \frac{QK^T}{\sqrt{d_k}} \,,
\end{equation}
where the scaling value of $d_k$ is the dimension of projected superpixel features. These attention weights are then used to scale the input features following 
\begin{equation}
    \text{head}(\bar{\mathcal{R}_k}) = \text{softmax}(\alpha) V \,.
\end{equation}
For multi-head self-attention, the results of multiple heads are concatenated together (denoted with $[\cdot \| \cdot]$) and reprojected with a learnt weight matrix $W_O$
\begin{equation}
    \text{Multihead}(Q, K, V) = [head_1 \| ... \| head_n]W_O
\end{equation}
Finally, the latent features are produced with a feed-forward network
\begin{equation}
    \bar{\mathcal{R}_k}' = \sigma(W_1 \mathbf{\bar{\mathcal{R}_k}})W_2 \,,
\end{equation}
where $\sigma$ is the ReLU non-linearity and $W_1$, $W_2$ are both learnable weight matrices. The encoder consists of a stack of multi-head self-attention blocks followed by a feed-forward network. The superpixel features $\mathcal{R} = \{\mathbf{r}_i\}^M_{i=0}$ are passed through the self-attention mechanism to calculate their corresponding attention weights. 

Once done for each resolution, the attended features are concatenated together to produce the final set of image features, $\mathbf{R}$ that will be used by the decoder to produce the caption.
\begin{equation}
    \mathbf{R} = [\bar{\mathcal{R}_1}' \| .. \| \bar{\mathcal{R}_k}']
\end{equation}
where $[\cdot \| \cdot]$ denotes concatenation. 

The decoder component of SuperCap then uses $\mathbf{R}$ alongside the previously generated word $w_{t-1}$ to autoregressively predict the next most likely word $w_t$.

\section{Implementation}
\label{sec:implementation}
In the interest of reproducibility, in section we detail our implementation choices as well as the training and evaluation protocol used. Our code is publicly available \footnote{Region Extraction: \url{https://anonymous.4open.science/r/superpixel-features-1754/}} \footnote{Captioning code: \url{https://anonymous.4open.science/r/DEFIGNN-B91F/README.md}}

\subsection{Model}
Our region extractor ($\varphi$) consists of a SLIC~\cite{achanta2010slic} superpixel segmentation process that computes $M$ superpixels from an image $\mathcal{I}$. Each superpixel is then converted into an $N$-dimensional feature vector by our region encoder ($\Psi$), which is implemented via either BLIP~\cite{li2022blip} ($N=768$) or CLIP~\cite{radford2021learning} ($N=512$). We demonstrate the effectiveness of both models in Section \ref{sec:results}. Once the superpixel features $\mathcal{R} = \{\mathbf{r}_i \in \mathbb{R}^N\}^M_{i=1}$ are computed, an additional global feature $\mathbf{r}_0 \in \mathbb{R}^N$ is extracted using the same feature extractor. Our captioning model ($\Omega$) is implemented via a six layer, eight head encoder-decoder Transformer. It makes use of sinusoidal positional encoding and predicts the next most likely token across a one-hot encoded dictionary of $10,000$ words. The dictionary is generated following the standard practice in image captioning~\cite{karpathy2015deep} of using words that appear more than five times in the corpus of captions.

\begin{table}[t!]
    \caption{\textbf{Comparison of Superpixel Resolutions}. We compare the impact of dividing the image into different numbers of superpixels using CLIP as the feature encoder. Results are for the Karpathy Test Split ~\cite{karpathy2015deep} of the COCO dataset~\cite{lin2014microsoft, chen2015microsoft}, where B-4, M, R, C, S denote BLEU@4, METEOR, ROUGE-L, CIDEr and SPICE scores respectively. Best results in \textbf{bold}.}
    \label{tab:clip-res}
    \centering
    \begin{tabular}{ccccccccc}
    \toprule
        \# Superpixels & B-4 $\uparrow$ & M $\uparrow$ & R $\uparrow$ & C $\uparrow$ & S $\uparrow$ \\
        \midrule
        1 & 35.0 & 27.2 & 56.4 & 119.1 & 20.6 \\
        2 & 21.5 & 20.1 & 46.0 & 71.4 & 13.4 \\
        3 & 25.5 & 22.4 & 49.4 & 88.0 & 15.9 \\
        4 & 34.7 & 27.2 & 56.4 & 118.4 & 20.7 \\
        5 & 37.5 & 28.4 & 58.1 & 127.1 & 22.1 \\
        6 & 36.0 & 27.7 & 57.0 & 122.5 & 21.3 \\
        7 & 36.5 & 27.9 & 57.3 & 122.7 & 21.4 \\
        8 & 36.3 & 27.9 & 57.3 & 123.2 & 21.5 \\
        9 & 37.1 & 28.1 & 57.6 & 125.4 & 21.8 \\
        10 & 37.9 & 28.6 & \textbf{58.2} & \textbf{128.4} & 22.4 \\
        15 & \textbf{38.0} & \textbf{28.7} & \textbf{58.2} & \textbf{128.4} & \textbf{22.5} \\
        20 & 37.8 & 28.5 & 57.9 & 127.0 & 22.3 \\
        25 & 37.2 & 28.4 & 57.8 & 126.58 & 22.3 \\
        50 & 36.0 & 27.9 & 56.9 & 121.7 & 21.5 \\
        75 & 35.4 & 27.6 & 56.3 & 119.6 & 21.5 \\
        100 & 34.5 & 27.1 & 55.7 & 115.9 & 20.9 \\
    \bottomrule
    \end{tabular}
\end{table}

\subsection{Dataset}
The Microsoft Common Objects in Context (COCO) dataset~\cite{lin2014microsoft, chen2015microsoft} is a widely used dataset for image captioning. While the original release incorporated an evaluation server with a hidden test set, the community has unanimously adopted what has become known as the Karpathy Split ~\cite{karpathy2015deep}. This is a train/val/test split of: $113,287$ / $5000$ / $5000$. Each image is accompanied by five human generated captions, some with differing focuses or verbosity. Overall, it provides a wide range of real life imagery alongside high quality captions.

\subsection{Training}
We train our model for $30$ epochs using the Adam~\cite{kingma2014adam} optimiser with the same learning rate scheduling strategy as ~\cite{cornia2020meshed,vaswani2017attention} during the cross-entropy loss phrase. Following standard practices in image captioning~\cite{fang2022injecting, yao2018exploring, zhong2020comprehensive, zhao2019object, rennie2017self}, we then use the CIDEr-based reinforcement learning strategy to prevent our model suffering from the exposure bias problem. The finetuning is completed over an additional 30 epochs, optimised by Adam~\cite{kingma2014adam} with a fixed learning rate of $5e-6$.

\begin{table}[t!]
    \caption{\textbf{Comparison of Superpixel Resolutions with Global Feature}. We compare the impact of dividing the image into different numbers of superpixels using CLIP as the feature encoder alongside the inclusion of a global image feature. Results are for the Karpathy Test Split ~\cite{karpathy2015deep} of the COCO dataset~\cite{lin2014microsoft, chen2015microsoft}, where B-4, M, R, C, S denote BLEU@4, METEOR, ROUGE-L, CIDEr and SPICE scores respectively. Best results in \textbf{bold}.}
    \label{tab:global-features}
    \centering
    \begin{tabular}{ccccccccc}
    \toprule
        \# Superpixels & B-4 $\uparrow$ & M $\uparrow$ & R $\uparrow$ & C $\uparrow$ & S $\uparrow$ \\ \midrule
        1 & 35.0 & 27.2 & 56.4 & 119.1 & 20.6 \\
        10 & \textbf{38.1} & 28.6 & 58.4 & \textbf{129.9} & 22.4 \\ 
        15 & 38.0 & \textbf{28.7} & 58.4 & 129.2 & \textbf{22.5} \\ 
        25 & 38.0 & \textbf{28.7} & \textbf{58.5} & 129.8 & 22.4 \\ 
        50 & 37.9 & 28.6 & 58.3 & 129.1 & 22.2 \\ 
        75 & 36.9 & 28.3 & 57.9 & 126.0 & 21.9 \\ 
        100 & 36.3 & 27.8 & 57.1 & 124.4 & 21.5 \\ 
        \bottomrule
    \end{tabular}
\end{table}

\subsection{Evaluation}
In line with other works in the field~\cite{fang2022injecting, yao2018exploring, zhong2020comprehensive, zhao2019object, rennie2017self}, we use a number of evaluation metrics~\cite{papineni2002bleu, lin2004rouge, banerjee2005meteor, vedantam2015cider, anderson2016spice}.

BLEU@$N$~\cite{papineni2002bleu} is a precision-based n-gram matching metric originally adopted from machine translation. It has long been used in image captioning and is one of the key evaluation metrics. Additionally from natural language processing, ROUGE~\cite{lin2004rouge} is a set of metrics used to evaluate the quality of automatic summarisation and machine translation systems. It takes into account: n-gram recall between the candidate and reference set, a comparison of the longest common sub-sequence, a comparison of the weighted longest common sub-sequence, and finally, the skip-bigram co-occurrence statistic. The final natural language processing evaluation used in image captioning is METEOR~\cite{banerjee2005meteor}, which uses the $F_{mean}$ to take into account both recall and precision. CIDEr~\cite{vedantam2015cider} and SPICE~\cite{anderson2016spice} are both evaluation metrics designed specifically for image captioning and have therefore become important benchmarks since their introduction.

Finally, note that during inference, we use a beam width of $5$ and a max decoding length of 20 words.

\section{Results}\label{sec:results}
Given the setting described in the previous Section, we perform a number of experiments aimed at evaluating the impact of our architectural choices on model performance as well as comparing SuperCap with state-of-the-art (SOTA) alternatives. We first compare the impact of various superpixel resolutions with the model running in a single resolution capacity. After seeing the strong performance of a single superpixel feature (essentially a global feature), we evaluate the impact of including a global superpixel feature into the architecture. Next we compare the impact of different region extractors and region encoders, comparing CLIP and BLIP superpixel features with BLIP patch features. Finally, we compare four different multi-resolution approaches before assembling our final model architecture to perform comparison against SOTA models. 

\subsection{Impact of Different Superpixel Resolutions}
Using CLIP to encode the superpixel region features, we compare the the performance of our single resolution model when changing the number of superpixels that the image is divided into (shown in Table  \ref{tab:clip-res}). We observe that whilst a single superpixel performs well (as it is essentially just a global context feature), there is a sweet spot in performance around 10 to 15 superpixels per image. Increasing the number of superpixels results in a decreasing performance. We hypothesise that this is because the superpixels are too small to contain enough of the image for the VLM to produce a meaningful feature.

\begin{table}[t]
    \caption{\textbf{Comparison of Different Initial Encodings}. We compare the impact of various initial encodings with all ablations including an appropriate global feature. Results are for the Karpathy Test Split ~\cite{karpathy2015deep} of the COCO dataset~\cite{lin2014microsoft, chen2015microsoft}, where B-4, M, R, C, S denote BLEU@4, METEOR, ROUGE-L, CIDEr and SPICE scores respectively. Best results in \textbf{bold}.}
    \label{tab:initial-encodings}
    \centering
    \begin{tabular}{lcccccccc}
    \toprule
        Encoding & B-4 $\uparrow$ & M $\uparrow$ & R $\uparrow$ & C $\uparrow$ & S $\uparrow$ \\ \midrule
        15 SPs \& CLIP & 38.0 & 28.7 & 58.4 & 129.2 & 22.5 \\
        15 SPs \& BLIP & \textbf{40.1} & \textbf{29.6} & \textbf{59.6} & \textbf{135.7} & \textbf{23.3} \\ 
        16 Patches \& BLIP & 39.2 & 29.2 & 59.2 & 133.2 & 23.0 \\ 
        \bottomrule
    \end{tabular}
\end{table}

\subsection{Impact of Global Features}
Using CLIP to encode the superpixel region features alongside a global image feature, we compare the the performance of our single resolution model when changing the number of superpixels that the image is divided into (shown in Table  \ref{tab:global-features}). By including the global feature, all superpixel resolutions see an improvement in performance when compared with their counterparts in Table \ref{tab:clip-res}. Note also that while the impact of adding the global feature appears to be uneven across the range of superpixel resolution (\emph{i.e.}, it has a larger positive impact for higher resolutions), this only minimally affects the optimum.

\begin{table}[t!]
    \caption{\textbf{Comparison of Different Multi-Resolution Approaches}. We compare the impact of various multi-resolution approaches alongside the impact of the inclusion of a global image feature $\mathbf{r}_0$ (denoted with FC in the table). Results are for the Karpathy Test Split ~\cite{karpathy2015deep} of the COCO dataset~\cite{lin2014microsoft, chen2015microsoft}, where B-4, M, R, C, S denote BLEU@4, METEOR, ROUGE-L, CIDEr and SPICE scores respectively. Best results in \textbf{bold}.}
    \label{tab:multi-res}
    \centering
    \begin{tabular}{lcccccccc}
    \toprule
        Encoding & B-4 $\uparrow$ & M $\uparrow$ & R $\uparrow$ & C $\uparrow$ & S $\uparrow$ \\ \midrule
        Method 1 & 37.6 & 28.3 & 57.9 & 127.3 & 22.2 \\ 
        Method 1 \& FC & 37.9 & 28.7 & 58.4 & 129.5 & 22.5 \\
        
        Method 2 & 38.5 & 28.8 & 58.7 & 129.9 & 22.7 \\ 
        Method 2 \& FC & 38.8 & 29.1 & 59.0 & 131.8 & \textbf{22.8} \\
        
        Method 3 & 36.4 & 27.8 & 57.2 & 123.4 & 21.2 \\ 
        Method 3 \& FC & 37.5 & 28.3 & 58.1 & 127.2 & 21.8 \\ 
        
        Method 4 & 37.4 & 28.2 & 57.8 & 125.9 & 22.0 \\ 
        Method 4 \& FC & 36.0 & 27.5 & 57.3 & 121.6 & 20.9 \\
        
        \midrule
        Dual \& FC & \textbf{39.2} & \textbf{29.2} & \textbf{59.1} & \textbf{132.4} & \textbf{22.8} \\
        
        \bottomrule
    \end{tabular}
\end{table}

\subsection{Comparison of Different Initial Encoders}
Here we explore the impact of different initial encoders by comparing both CLIP~\cite{radford2021learning} and BLIP~\cite{li2022blip} embeddings of the same SLIC~\cite{achanta2010slic} superpixels. As shown in Table \ref{tab:initial-encodings}, for the same superpixels, BLIP is superior in performance across all the benchmarks. We then hold the embedding model constant and change the region representation from superpixels to the more classic patches. We create $16$ patches by dividing the image into a $4 \times 4$ grid and produce BLIP features for each patch. Whilst traditional patching methods like those used in Vision Transformer~\cite{dosovitskiy2020image} use a smaller grid with more regions, we demonstrate in Tables \ref{tab:clip-res} and \ref{tab:global-features} that higher region counts correlate with lower performance. Therefore, in order to ensure a fair comparison we use a comparable amount of regions.

\begin{table*}[t!]
 \caption{\textbf{Comparison With SOTA}. The below table compares our SuperCap to the current SOTA models that perform detector-free image captioning without vision-language pretraining. Results are for the Karpathy Test Split ~\cite{karpathy2015deep} of the COCO dataset~\cite{lin2014microsoft, chen2015microsoft}. End-to-end pretrained models are denoted with a *.
 Best scores for \emph{`detector'} based models are indicated in \emph{italics} and the best scores for \textbf{detector-free} models are indicated in \textbf{bold}.}
 \label{tab:results}
\centering
\begin{tabular}{lccccccccccc}
\toprule
Model & \# Trainable Params $\downarrow$ & BLEU-4 $\uparrow$ & METEOR $\uparrow$ & ROUGE-L $\uparrow$ & CIDEr $\uparrow$ & SPICE $\uparrow$ \\ \midrule
Detector Models & & & & & & \\
RFNet~\cite{jiang2018recurrent}     & - & 36.5 & 27.7 & 57.3 & 121.9 & 21.2 \\
BUTD~\cite{anderson2018bottom}      & - & 36.3 & 27.7 & 56.9 & 120.1 & 21.4 \\
SGAE~\cite{yang2019auto}            & - & 38.4 & 28.4 & 58.6 & 127.8 & 22.1 \\
AoANet~\cite{huang2019attention}    & - & 38.9 & 29.2 & 58.8 & 129.8 & 22.4 \\
M2~\cite{cornia2020meshed}          & - & 39.1 & 29.2 & 58.6 & 131.2 & 22.6 \\
X-LAN~\cite{pan2020x}               & - & 39.5 & 29.5 & 59.2 & 132.0 & 23.4 \\
RSTNet~\cite{zhang2021rstnet}       & - & 40.1 & 29.8 & \textit{59.5} & 135.6 & 23.3 \\
OSCAR$_b$*~\cite{li2020oscar}       & \textit{198M} & 40.5 & 29.7 & - & 137.6 & 22.8 \\
VinVL$_b$*~\cite{zhang2021vinvl}    & 260M & \textit{40.9} & \textit{30.9} & - & \textit{140.4} & \textit{25.1} \\
\midrule
Detector Free Models & & & & & & \\
CLIP-CAP~\cite{mokady2021clipcap}   & - & 33.5 & 27.5 & -    & 113.1 & 21.05 \\
BLIP* ~\cite{li2022blip}            & - & \textbf{40.4} & -    & -    & 136.7 & -    \\
ViTCap~\cite{fang2022injecting}     & 190M & 40.1 & 29.4 & 59.4 & 133.1 & 23.0 \\
SuperCap CLIP (Ours)              & \textbf{45M} & 39.2 & 29.2 & 59.1 & 132.4 & 22.8 \\ 
SuperCap BLIP (Ours)              & 93M & 40.0 & \textbf{29.8} & \textbf{59.7} & \textbf{136.9} & \textbf{23.6} \\
\bottomrule
\end{tabular}
\end{table*}

\subsection{Different Multi-Resolution Approaches}
Extending our architecture to support multiple resolutions can be achieved in a variety of different ways. \textbf{Method 1} is the most straightforward approach, and involves keeping the architecture the same and simply concatenating the different resolutions together. This essentially increased the number of tokens that our model receives. \textbf{Method 2}, which proved to be the most effective, involves stacking the secondary encoder, one per resolution. The outputs of these encoders are then concatenated and fed into a single decoder language model. \textbf{Method 3} and \textbf{Method 4} investigate using a mixture of experts (MoE) approach to the decoder, following previous captioning work~\cite{yang-etal-2023-transforming}. \textbf{Method 3} passes the resolutions through a single encoder with each resolution going to its own decoder. The outputs of the decoders are then passed through a soft routing network to select the final token. \textbf{Method 4} follows the same MoE approach but uses one encoder per resolution. 

We test each of the four methods with CLIP~\cite{radford2021learning} superpixels generated at four resolutions, $25$, $50$, $75$, $100$ to ensure that the models receive inputs that cover a wide range of scale. Our ablation concludes by testing the best multi-resolution approach (\textbf{Method 2}) with dual inputs (referred to in Table \ref{tab:multi-res} as `dual') $10$ and $25$ superpixels. 

\begin{figure*}[!htb]
\centering
\begin{tabular}{P{4cm}P{0cm}P{4cm}P{0cm}P{4cm}}
\includegraphics[width=0.25\textwidth, height=0.15\textheight]{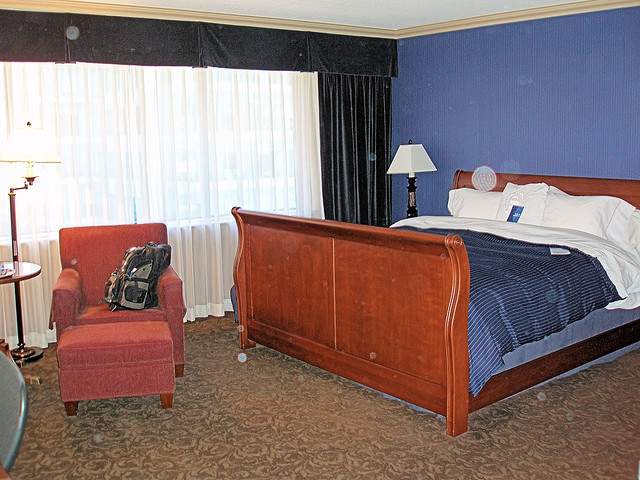} & & \includegraphics[width=0.25\textwidth, height=0.15\textheight]{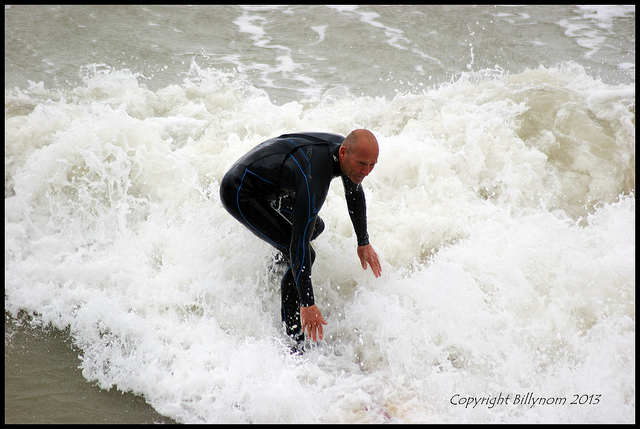} & & \includegraphics[width=0.25\textwidth, height=0.15\textheight]{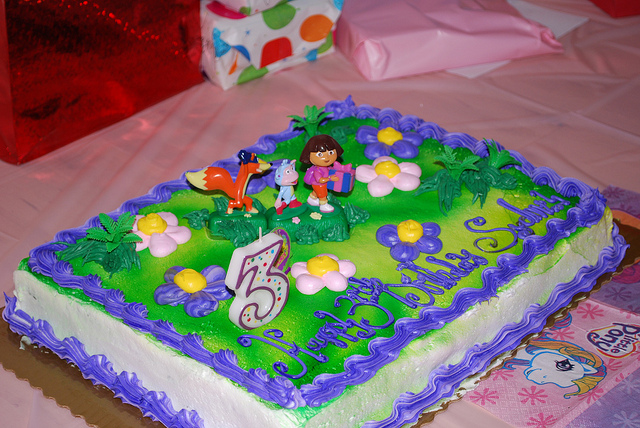}\\
Ours: \emph{a bedroom with a bed and a chair} & & Ours: \emph{a man riding a wave on a surfboard in the water} & & Ours: \emph{a cake with a train on top of a table} \\
GT: \emph{A view of a bed, chair, and window treatments.} & & GT: \emph{A man riding a wave on top of a surfboard.} & & GT: \emph{A cake with icing and cartoon cake toppers.} \\
\includegraphics[width=0.25\textwidth, height=0.15\textheight]{figs/test/COCO_val2014_000000081988.jpg} & & \includegraphics[width=0.25\textwidth, height=0.15\textheight]{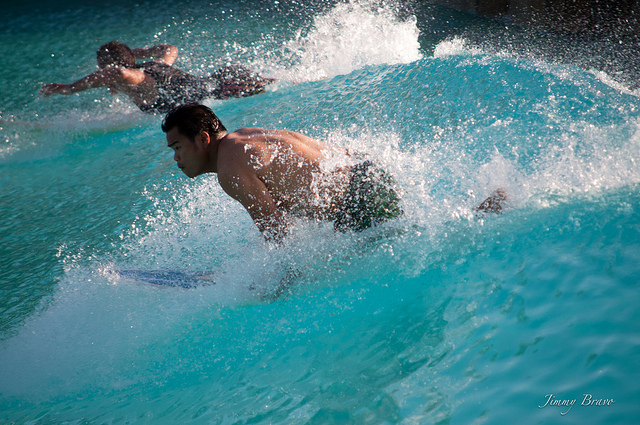} & & \includegraphics[width=0.25\textwidth, height=0.15\textheight]{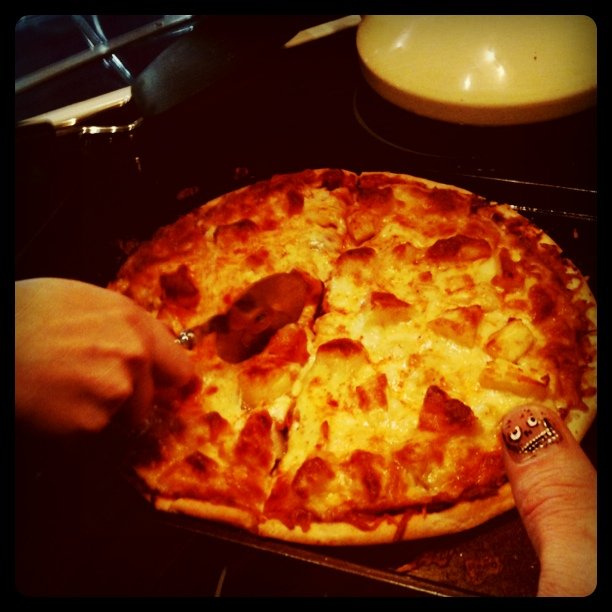} \\
Ours: \emph{a group of people on the beach with surfboards} & & Ours: \emph{a man riding a wave on a surfboard in the water} & & Ours: \emph{a person holding a pizza in front of a pan
} \\
GT: \emph{Four kids sitting on surfboards with man in background.} & & GT: \emph{Two people surfing a beautiful sky blue wave.} & & GT: \emph{A pizza is being cut into on a sheet.} \\
\end{tabular}
\caption{\textbf{Test Set Performance.} Images were randomly selected from the test set and captioned with the best performing model settings using beam width $5$. Best viewed in colour.}
\label{tab:qual-test-set}
\end{figure*}

\subsection{Comparison with SOTA}
In order to demonstrate the performance of our model, we compare it against similar detector-free image captioning models. To the best of our knowledge, as this is a new field, there only exists~\cite{fang2022injecting}. We therefore include some common detector-based image captioning models in order to place our work within a wider context.

When compared against the next best model~\cite{fang2022injecting}, our model has considerably fewer parameters. As shown in \Cref{tab:results}, we match or exceed the performance during the important CIDEr optimisation stage. This optimisation stage was introduced to reduce the exposure bias problem caused by training language models with teacher forcing~\cite{rennie2017self}, which is used during the cross-entropy loss training. Given the significant reduction in parameter count, alongside the strong performance, it is clear that our architecture is very parameter efficient.

\subsection{Qualitative Results}
In order to demonstrate our model in action, we present a representative selection of five randomly selected Karpathy test set \cite{karpathy2015deep} images, accompanied by our generated model and their ground truth captions. Our model produces both accurate and descriptive captions and demonstrates an ability to identify individual objects (such as the chair and bed in the top left image). Even in cases where the caption does not perfectly match the ground truth, such as in the bottom right picture, the caption is nevertheless an accurate description of the image. 

Like all image captioning models, SuperCap does also suffer from mild hallucination, as in the case of the top right image. Our model describes a `train' being present on top of the cake, whereas in actuality it is a children's character. Another possible shortcoming present in Figure \ref{tab:qual-test-set} is the production of identical captions for similar images, as in the case for the two central images.

\section{Conclusion}
\label{sec:conclusion}
In this paper we introduced SuperCap, a superpixel-based image captioning architecture with support for multi-resolution. Our architecture makes use of the zero shot classification capabilities of large pretrained models, meaning that it is easily scalable to account for advancements in these large models. We hope that our work demonstrates to the community that irregular groups of homogeneous pixels can be used instead of the traditional object bounding boxes or patches. Future work will investigate alternative ways to handle multiple resolutions. 

{
    \small
    \bibliographystyle{ieeenat_fullname}
    \bibliography{main}
}

\clearpage
\setcounter{page}{1}
\maketitlesupplementary


\section{Further Qualitative Results}
\begin{figure*}[b]
\centering
\begin{tabular}{P{4cm}P{0cm}P{4cm}P{0cm}P{4cm}}
\includegraphics[width=0.25\textwidth, height=0.15\textheight]{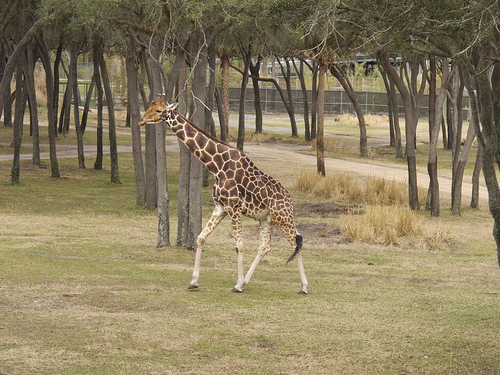} & & \includegraphics[width=0.25\textwidth, height=0.15\textheight]{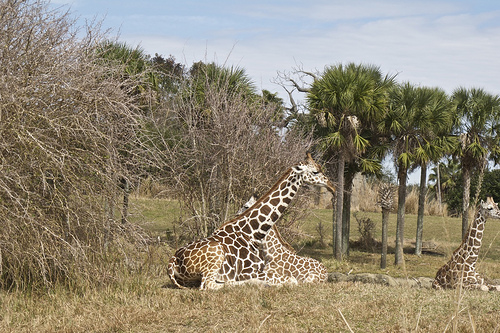} & & \includegraphics[width=0.25\textwidth, height=0.15\textheight]{sm/COCO_val2014_000000341921.jpg}\\
Ours: \emph{a giraffe walking in the grass in a field} & & Ours: \emph{two giraffes and a giraffe laying in a field} & & Ours: \emph{a young boy sitting on a bench in a park} \\
GT: \emph{a giraffe walks on the tundra tree-lined park} & & GT: \emph{Three giraffes are sitting on the ground} & & GT: \emph{A young man sitting on a park bench next to a playground} \\

\includegraphics[width=0.25\textwidth, height=0.15\textheight]{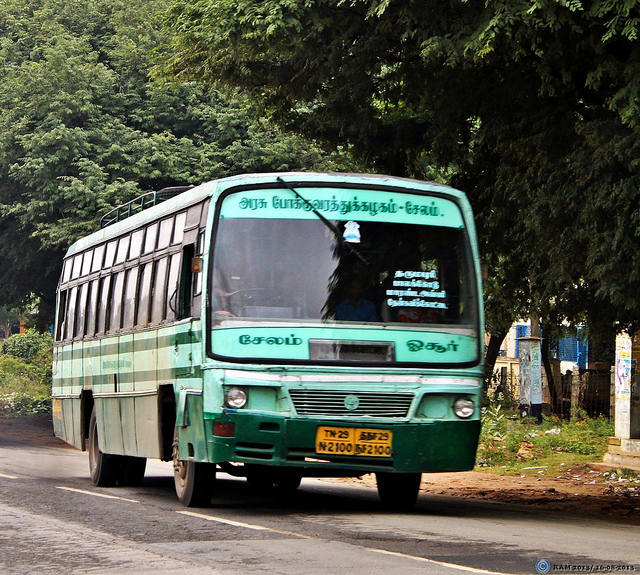} & & \includegraphics[width=0.25\textwidth, height=0.15\textheight]{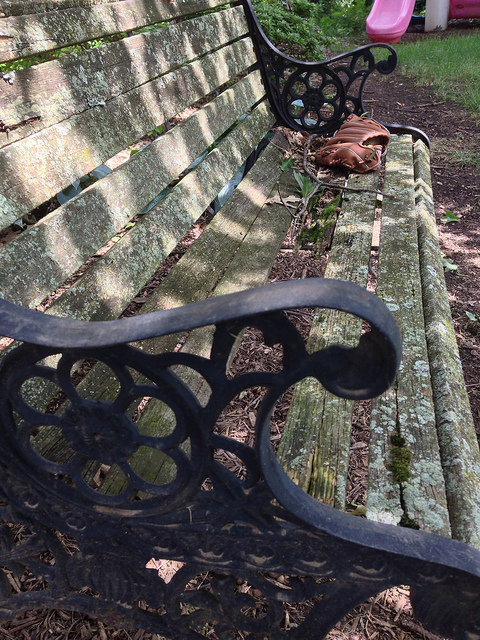} & & \includegraphics[width=0.25\textwidth, height=0.15\textheight]{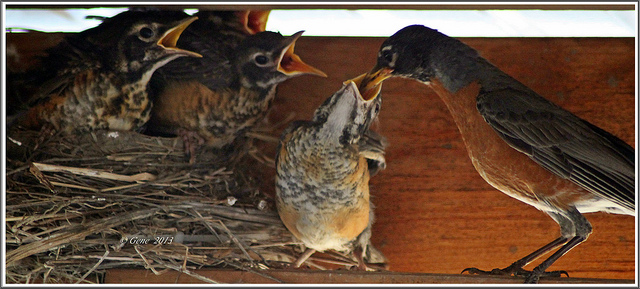} \\
Ours: \emph{a green bus driving down a street} & & Ours: \emph{a wooden bench sitting on top of a park} & & Ours: \emph{a group of birds eating from a nest} \\
GT: \emph{a green bus is driving down the street} & & GT: \emph{A closeup of an old and mossy outdoor garden bench} & & GT: \emph{A mother bird feeds her little baby chicks} \\

\includegraphics[width=0.25\textwidth, height=0.15\textheight]{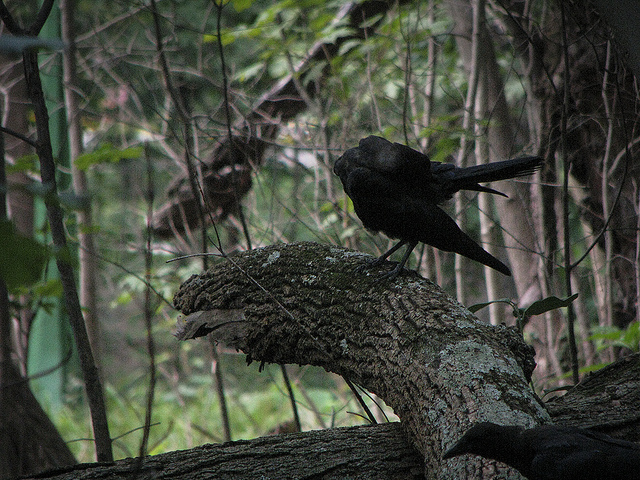} & & \includegraphics[width=0.25\textwidth, height=0.15\textheight]{sm/COCO_val2014_000000236690.jpg} & & \includegraphics[width=0.25\textwidth, height=0.15\textheight]{sm/COCO_val2014_000000423337.jpg} \\
Ours: \emph{a black bird perched on top of a tree branch} & & Ours: \emph{a white bird flying over a body of water} & & Ours: \emph{a cat laying on a bed next to a book} \\
GT: \emph{A couple of black birds are standing on a tree branch} & & GT: \emph{A white and gray bird soaring over the blue ocean} & & GT: \emph{A cat sleeping on a pillow next to a book} \\

\includegraphics[width=0.25\textwidth, height=0.15\textheight]{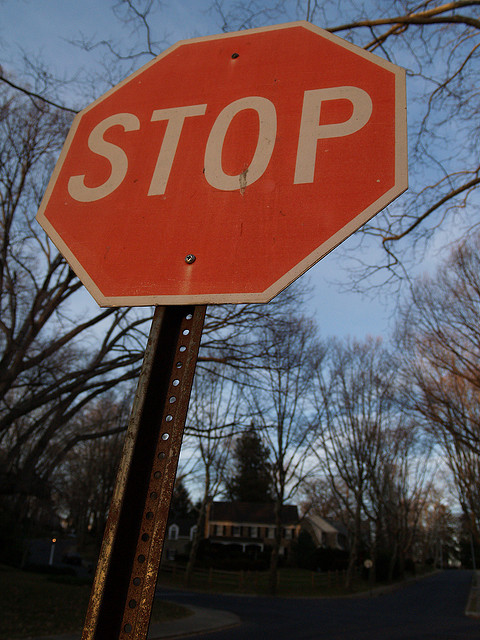} & & \includegraphics[width=0.25\textwidth, height=0.15\textheight]{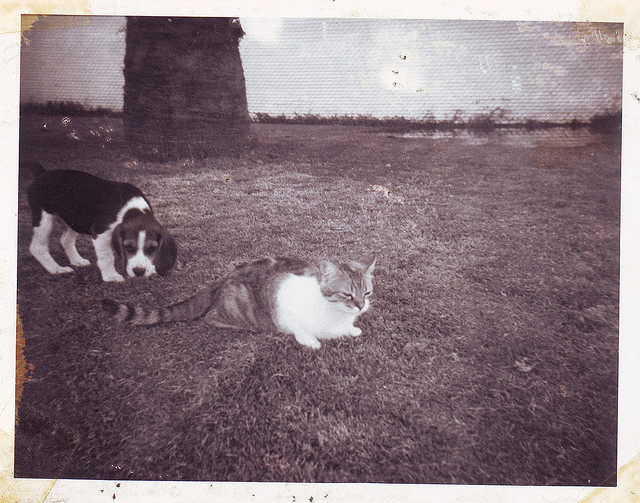} & & \includegraphics[width=0.25\textwidth, height=0.15\textheight]{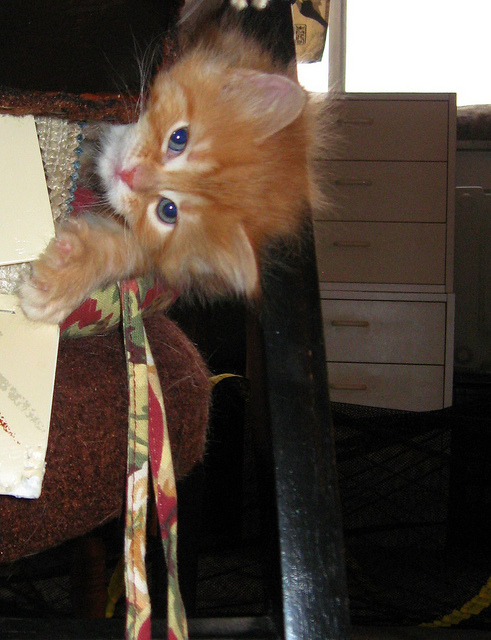} \\
Ours: \emph{a stop sign on the side of a street} & & Ours: \emph{a cat and a dog sitting in the grass} & & Ours: \emph{a cat sitting on top of a chair with a tie} \\
GT: \emph{A stop sign is viewed from a low angle} & & GT: \emph{A dog standing next to a cat in a  dirt field} & & GT: \emph{A kitten peeking out from behind a tablecloth on a chair} \\
\end{tabular}
\caption{\textbf{Qualitative Performance.} Example captions from the Karpathy test set. Best viewed in colour.}
\label{sm:qual-test-set}
\end{figure*}

\end{document}